\documentclass[conference]{IEEEtran}
\IEEEoverridecommandlockouts
\usepackage{cite}
\usepackage{amsmath,amssymb,amsfonts,algorithm,algpseudocode}

\usepackage{graphicx}
\usepackage{textcomp}
\usepackage{xcolor}
\usepackage{bbm}
\usepackage{enumitem}
\usepackage{mathtools}

\def\BibTeX{{\rm B\kern-.05em{\sc i\kern-.025em b}\kern-.08em
    T\kern-.1667em\lower.7ex\hbox{E}\kern-.125emX}}
\begin{document}

\title{Click-Based Student Performance Prediction: \\ A Clustering Guided Meta-Learning Approach
}
\author{\IEEEauthorblockN{Yun-Wei Chu$^1$, Elizabeth Tenorio$^1$, Laura Cruz$^1$, Kerrie Douglas$^1$, Andrew S. Lan$^2$, Christopher G. Brinton$^1$ \\
\IEEEauthorblockA{$^1$Purdue University, $^2$University of Massachusetts Amherst}
$^1$\{chu198, etenori, lcruzcas, douglask, cgb\}@purdue.edu, $^2$andrewlan@cs.umass.edu
}}


\maketitle

\newcommand{\yun}[1]{{\small\color{magenta}{\bf\xspace#1 -yun}}}

\begin{abstract}
We study the problem of predicting student knowledge acquisition in online courses from clickstream behavior. Motivated by the proliferation of eLearning lecture delivery, we specifically focus on student in-video activity in lectures videos, which consist of content and in-video quizzes. Our methodology for predicting in-video quiz performance is based on three key ideas we develop. First, we model students' clicking behavior via time-series learning architectures operating on raw event data, rather than defining hand-crafted features as in existing approaches that may lose important information embedded within the click sequences. Second, we develop a \emph{self-supervised clickstream pre-training} to learn informative representations of clickstream events that can initialize the prediction model effectively.
Third, we propose a \emph{clustering guided meta-learning-based training} that optimizes the prediction model to exploit clusters of frequent patterns in student clickstream sequences.
Through experiments on three real-world datasets, we demonstrate that our method obtains substantial improvements over two baseline models in predicting students' in-video quiz performance. Further, we validate the importance of the pre-training and meta-learning components of our framework through ablation studies. Finally, we show how our methodology reveals insights on video-watching behavior associated with knowledge acquisition for useful learning analytics.
\end{abstract}


\section{Introduction}
The proliferation of electronic learning (eLearning) platforms such as Coursera and edX have enabled worldwide student access to online course content~\cite{Martin2012WillMO, Wilkowski2014StudentSA, Guo2014DemographicDI, Zhang2017DynamicKM}.
Most recently, eLearning has proven essential due to the COVID-19 pandemic, during which the number of online learners skyrocketed~\cite{Adedoyin2020Covid19PA}.
The unfamiliarity of traditional classroom students with online education has brought the issue of quality control to the forefront, identifying a need to increase the effectiveness of online learning~\cite{doyumgacc2021understanding}.

Contemporary eLearning platforms collect a substantial amount of data on student interactions. This brings novel opportunities to study the process of human learning, and in turn to improve user experience on eLearning platforms, e.g., through predictive learning analytics and content personalization~\cite{Brinton2014LearningAS, Brinton2016MiningMC, Shridharan2018PredictiveLA}. Specifically, learning management systems employed at educational institutions are typically capable of collecting data on quiz/assessment responses, clickstream actions on user navigation through the platform, content access logs, social networking on discussion forums, and video watching behavior. Additional educational technologies have also been deployed at a smaller scale to capture certain types of student learning data, e.g., using cameras to capture facial expressions showing confusion and fatigue~\cite{Whitehill2014TheFO, Yang2016ExploringTE}.

\begin{figure*}[t]
\centering
\includegraphics[width=\linewidth]{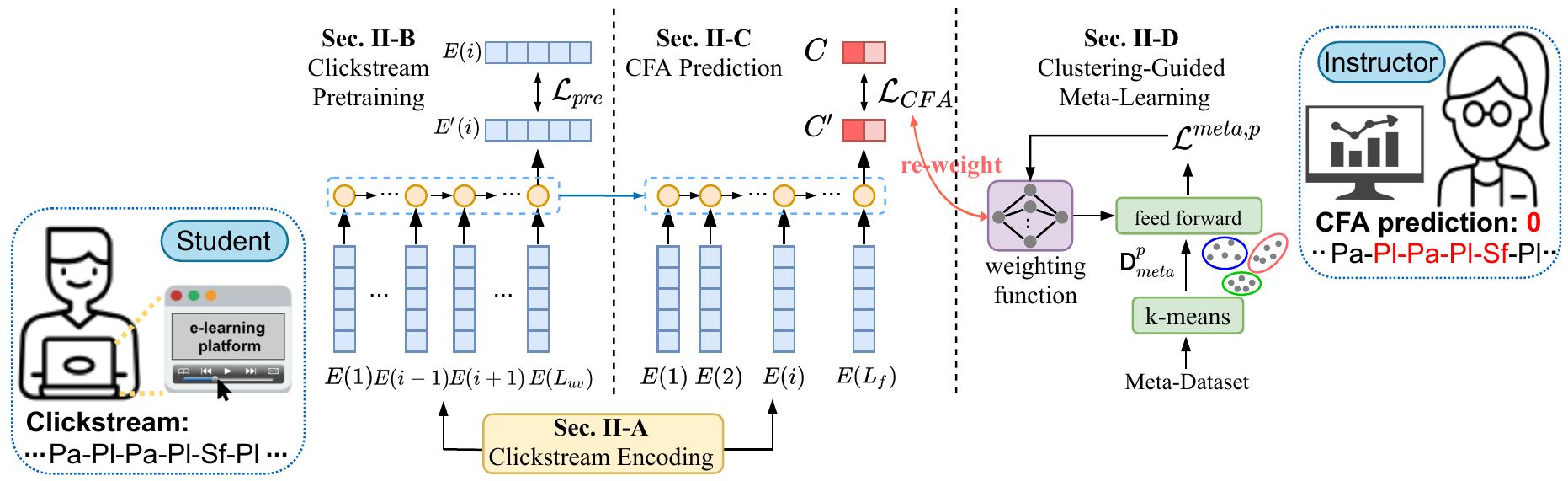}
\vspace{-0.3in}
\caption{Overview of the methodology we develop in this paper for predicting student in-video quiz performance based on video-watching click measurements. After encoding the clickstream data (Sec.~\ref{sec:extr}), the model first learns a surrogate objective by predicting the event type in a self-supervised manner (Sec.~\ref{sec:pre}). Based on this pre-training knowledge, a time-series neural network models students' click behavior from their raw clickstreams and predicts whether they are Correct on First Attempt (CFA) or not at the corresponding in-video question (Sec.~\ref{sec:cfa}). Finally, a meta-learning technique guides the model to incorporate information from student behavioral clusters during an optimization process to improve the performance of CFA prediction (Sec.~\ref{sec:optim}). 
}
\vspace{-0.2in}
\label{fig:overview}
\end{figure*}

The analysis of clickstreams generated by students as they navigate through online course materials has received considerable attention, even before COVID-19. It has been shown that the fine-granular nature of this data can facilitate the development of predictive models for learner knowledge transfer~\cite{Chan2021ClickstreamKT}, attrition/dropout likelihood~\cite{Kloft2014PredictingMD, Gitinabard2018YourAO}, and social engagement with classmates~\cite{Pendry2015IndividualAS}. With the growing trend of delivering courses online, one type of clickstream data that is becoming ubiquitous is learner interaction with \textit{lecture videos}. A considerable number of studies conducted on Massive Open Online Courses (MOOCs)~\cite{Sinha2014YourCD, Kloft2014PredictingMD, Nagrecha2017MOOCDP, Jeon2020DropoutPO} have shown that lecture video-watching clickstreams can be used to predict student engagement and dropout tendencies, which can provide useful analytics for instructors attempting to improve engagement in their courses.

On the other hand, relatively few works~\cite{Lan2017BehaviorBasedLV, Yang2018LearningIA, Yang2017BehaviorBasedGP} have considered how these clickstreams may be informative of student \textit{knowledge acuisition}. eLearning lecture videos are often accompanied by in-video quiz questions, which give a rapid indication of student performance. This presents an opportunity to model students' in-video behavior together with their quiz performance, which can provide insights into student learning and content efficacy~\cite{Halawa2014DropoutPI, Whitehill2015BeyondPT}, motivating our research in this work.

Most existing models leveraging student video-watching clickstream behavior for prediction tasks have relied on \textit{hand-crafted features}, where the clickstreams are pre-processed into summary quantities (e.g., time spent, number of plays/pauses, etc.). These features are then used to train either standard machine learning models~\cite{Kloft2014PredictingMD, Taylor2014LikelyTS, Brinton2015MOOCPP} or deep learning models~\cite{Wang2017DeepMF, Fei2015TemporalMF} to predict engagement or performance. However, e process of hand-crafting my discard valuable information within the clickstreams, such as their sequential patterns~\cite{Yang2018LearningIA, Jeon2020DropoutPO_1}. We are therefore motivated to investigate modeling students' clicking behavior from their \textit{raw clickstreams}, i.e., without any feature engineering, to maintain the primitive information. In doing so, the key challenge is the inherent noisy nature of these clickstreams, e.g., students making accidental clicks that are not associated with their learning process~\cite{Brinton2016MiningMC}.
This motivates an investigation of \textit{self-supervised pre-training} for modeling the underlying clickstream generation process~\cite{Sinha2014YourCD}. Research on both in-person~\cite{ebi2020StudentsIP} and online~\cite{Zhang2020MOOCVP, Shi2020SocialIC, Boroujeni2018DiscoveryAT, Williams2018CharacterizingML, Tan2018LearningPB, Zhang2020MOOCVP} courses has also found that clusters of student learning behavior exist (e.g., corresponding to different learning strategies). 
Thus we are also motivated to investigate how learning clusters of student behaviors can enhance our video-watching performance prediction methodology. 

To summarize, we pose the following research questions:
\begin{enumerate}
\item \textit{How can we model the process of student lecture clicking behavior in the manner that is predictive of in-video quiz performance?}
\item \textit{Can inferred clustering of students' clicking behavior based on their observed interaction benefit the prediction significantly?}
\item \textit{How can our model reveal analytics on the relationship between video-watching behavior and quiz performance?}
\end{enumerate}

To answer these questions, we develop a novel deep learning-based methodology that predicts students' in-video quiz performance -- specifically, the likelihood that a student will be Correct on their First Attempt (CFA) at answering the question -- based on their clicking behavior. 
We focus on CFA because it can reduce confounding factors unrelated to the lecture video, e.g., reviewing other materials~\cite{Brinton2016MiningMC, Brinton2015MOOCPP, Jahrer2010CollaborativeFA}.
A key technique we will develop to improve the prediction quality is \textit{clustering-guided meta-learning}, which guides the neural network to reflect student behavioral clusters during the optimization process. 
Our evaluation on real-world datasets shows that the behavioral patterns extracted from this process provide useful learning analytics.

\subsection{Related Work}
\subsubsection{Student behavioral mining}
Measurements of student eLearning behavior can be \textit{device-based} or \textit{activity-based}. Device-based methods use devices external to the eLearning platform, e.g., cameras and eye-trackers, which have been shown to be indicative of learning metrics such as confusion and fatigue ~\cite{Whitehill2014TheFO, Dhall2020EmotiW2D, Thomas2018MultimodalTA,Yang2016ExploringTE}. In this work, we are focused instead on activity-based measurements from eLearning platforms, e.g., from discussion boards~\cite{lan2018personalized,brinton2018efficiency}. In particular, we consider video watching behavior, typically recorded as sequences of events such as pause, play, and skip made by students interacting with lecture video players~\cite{Yu2019PredictingLO, Luo2018AMV}. A few works have investigated sequential pattern mining techniques for extracting subsequences of clicks from lecture video-watching~\cite{Sinha2014YourCD,Brinton2016MiningMC,chen2018behavioral,cao2018learner}. Specifically,~\cite{Sinha2014YourCD} analyzed the most frequently occurring $n$-gram click actions from a contiguous sequences of clicks, while~\cite{Brinton2016MiningMC} applied probabilistic mixture modeling algorithms to extract recurring subsequences of events. We are instead interested in supervised approaches to predicting knowledge acquisition from clicks.

\subsubsection{Video-watching behavior and performance prediction}
eLearning courses often includes in-video quiz questions, serving as immediate feedback for both instructor and learners of the students' knowledge gains. It was shown in~\cite{Brinton2015MOOCPP} that certain attributes of video-watching behavior (e.g., fraction completed, average playback rate) are correlated with in-video quiz performance. Motivated by this, researchers have studied the potential of leveraging video-watching behavior for early detectors of performance on in-video questions. Specifically, ~\cite{Lan2017BehaviorBasedLV} proposed a two-stage probabilistic latent variable model to predict CFA performance from hand-crafted features identified in~\cite{Brinton2015MOOCPP} as being correlated with knowledge acquisition. In another study,~\cite{Brinton2016MiningMC} proposed a maximum likelihood estimation approach for CFA prediction based on visits to specific positions and transitions in lecture videos. Finally,~\cite{Yang2018LearningIA} employed Generative Adversarial Networks (GAN) to predict CFA from hand-crafted features while simultaneously safeguarding sensitive attributes within the data. Compared to these approaches, we develop techniques that remove the hand-crafted feature pre-processing step, which we will show enhances CFA prediction performance.




\subsubsection{Clustering and meta-learning}
Inspired by previous research~\cite{Zhang2020MOOCVP, Shi2020SocialIC, Boroujeni2018DiscoveryAT, Williams2018CharacterizingML, Tan2018LearningPB} that has shown the potential of clustering students’ learning behavior, we leverage meta-learning to optimize our prediction methodology based on student similarities within clusters. Meta-learning~\cite{Finn2017ModelAgnosticMF} is a branch of machine learning dealing with tasks requiring the machine learning to improve its learning process~\cite{jordan2015machine}; it has found success in educational applications for tasks such as grading~\cite{wang2019meta} and computerized adaptive testing~\cite{ghosh2021bobcat}. A significant amount of research in meta-learning has focused on solving label imbalances through sample reweighting approaches~\cite{Malisiewicz2011EnsembleOE, Lin2020FocalLF, Kumar2010SelfPacedLF}. The central concept of these reweighing approaches is to design a weighting function that maps from training loss to sample weights based on a set of pre-determined hyperparameters.
In our setting, instead of using a fixed weighting function, we consider MW-net~\cite{Shu2019MetaWeightNetLA}, a dynamic weighting function that automatically learns the hyperparameters for meta-learning optimization.
We will show that our meta-learning approach for training optimization based on student groups leads to significant improvements in CFA prediction quality.

\subsection{Overview of Methodology and Contributions}
In this paper, we develop a novel methodology for predicting in-video knowledge acquisition that considers the sequential clicking behavior of individual students and similarity between student clusters. An overview of our methodology is shown in Figure~\ref{fig:overview}.
After encoding the raw video-watching clickstreams generated from eLearning platforms (Section~\ref{sec:extr}), the first part of our framework is \textit{clickstream pre-training}, where we develop a self-supervised algorithm for modeling student click generation (Section~\ref{sec:pre}).
Specifically, we learn a denoised representation of each event using surrounding click events to predict the current click from every sequence.
The next part of our framework is \textit{CFA prediction}, where the self-supervised pre-training is leveraged by time-series deep learning in modeling users' sequential clickstreams for predicting their in-video quiz performance (Section~\ref{sec:cfa}). The CFA output is a binary value of success or failure on the in-video quiz question. In the third part of our framework, \textit{clustering-guided meta-learning}, we cluster a portion of the dataset based on students' clicking behavior to partition students into groups. Our meta-learning-based training procedure then brings similar information within each cluster to the CFA prediction model during the optimization process (Section~\ref{sec:optim}). The dynamic weighting function learns from each student cluster and guides the loss of the CFA prediction model to reflect the information of each group. 

Overall, in this work, we make the following contributions:
\begin{itemize}
\item We develop a novel methodology for predicting student knowledge acquisition based on video-watching click measurements collected by eLearning platforms. Our method eliminates the need in prior approaches for hand-crafted feature engineering through time-series deep learning models of the raw click events.

\item To enhance our prediction model's capability, we develop a self-supervised clickstream pre-training method to model student click generation. We also develop a meta-learning-based training procedure that guides the CFA prediction model to reflect inferred similarities within student behavioral clusters.

\item Our experiments (Section~\ref{sec:experiment}) on three real-world datasets reveal that our proposed methodology obtains substantial improvements in accuracy and F1 scores over existing CFA prediction algorithms. Our ablation studies confirm the benefit provided by the self-supervised pre-training and clustering-guided meta-learning components of our methodology.
Additionally, we show how the clickstream-to-CFA relationships learned by our method can provide useful learning and content analytics.


\end{itemize}


\section{Prediction Methodology}
\label{sec:method}
In this section, we introduce our clickstream-based CFA prediction methodology.
First, we present data encodings for video-watching click sequences and quiz performance (Section~\ref{sec:extr}). We then develop a self-supervised pre-training algorithm to model the process of students generating clicks (Section~\ref{sec:pre}). 
Our base CFA prediction model (Section~\ref{sec:cfa}) models students' clicks in a recurrent neural network (RNN)-based algorithm.
Finally, we develop a meta-learning-based training procedure that optimizes CFA learning based on inferred similarities between clusters of student behaviors (Section~\ref{sec:optim}).

\subsection{Data Processing and Clickstream Encodings}
\label{sec:extr}
We focus on two types of video interactions collected by eLearning platforms: (i) video-watching click actions and (ii) answers to in-video quiz questions. 
Later in Section~\ref{sec:experiment} we will implement our method on datasets collected from two platforms, edX and Coursera; the specifics of these datasets will be discussed in Section~\ref{sec:dataset}. 

\subsubsection{Video-watching clickstreams events}
Click events include playing, pausing, changing the playback speed, and skipping to another place in the video. Each time one of these events is fired, a data entry is recorded that specifies the user ID, video ID, event type, and UNIX timestamp for the event. All the students' video-watching behavior is recorded as a chronological sequence of click events.

Formally, we represent the $i$-th event made by user $u$ on video $v$ as $E_{uv}(i) = (e_i, p_i, t_i, s_i, r_i)$, where $i \in \{1, ..., L_{uv}\}$ and $L_{uv}$ denotes the total number of clicks user $u$ made on video $v$.
In each event $E_{uv}(i)$, $e_i$ is the type of the $i$-th click (defined below), $p_i$ is the video position of the player (in seconds), $t_i$ is the UNIX time at which event was fired, $s_i \in \{0,1\}$ is the binary playback state -- either playing or paused -- of the video player, and $r_i$ is the speed of the video player. 
$e_i \in \{0,...,4\}$ can be one of five types : 
\begin{enumerate}[label=(\roman*)]
\item \textsf{play} (Pl): A play event begins when a click event $E_{uv}(i)$ is made for which the state is playing ($s_i = 1$).
\item \textsf{pause} (Pa): A pause event stored when a click event $E_{uv}(i)$ is made for which the state is paused ($s_i = 0$).
\item \textsf{skip back} (Sb): A skip back event occurs when $p'_{i} > p_i$, where $p'_{i}$ is the position of the video player immediately before the skip.
If $s_{i-1} = 1$ (playing), $p'_i = p_{i-1} + r_{i-1} \cdot (t_i - t_{i-1}) $; if $s_{i-1} = 0$ (paused), then $p'_i = p_{i-1}$.
\item \textsf{skip forward} (Sf): A skip forward event is defined in the same way as skip back event, except it occurs when $p'_{i} < p_i$.
\item \textsf{rate/speed change} (Sp): A playback rate/speed change event occurs when $r'_{i} \neq r_{i}$, where  $r'_{i}$ is the speed of the video player immediately before the rate change.
\end{enumerate}
Instead of saving all the click events as a sequence, we combine repeated events that occur within a short duration (5 seconds) of one another since this pattern indicates that the user was adjusting to a final state. 
For each student-video $(u,v)$ pair, we refer to $E_{uv} = \{E_{uv}(i)\}$ as the resulting event sequence.
The distribution of each event type for our three datasets (discussed in Section~\ref{sec:dataset}) is listed in Table~\ref{tb:click}.
\begin{table}
\vspace{0.1in}
\begin{center}
\scalebox{0.86}{
 \begin{tabular}{lccccc}
\hline
& play & pause & skip back & skip forward & rate change\\
\hline\hline

edX-Purdue & 436,087 & 279,562 & 134,320 & 145,918 & 5,840\\
Coursera-FMB & 407,088 & 302,542 & 48,369 & 44,139 & 30,204\\
Coursera-NI & 332,581 & 251,719 & 30,885 & 26,237 & 112,411\\

\hline
\end{tabular}}
\end{center}
\caption{Distribution of each event type for our three video-watching clickstream datasets.}
\vspace{-0.25in}
\label{tb:click}
\end{table}

\subsubsection{In-video quiz performance}
We consider lecture videos that are equipped with an in-video quiz, e.g., multiple-choice or True/False questions. When a student $u$ submits an answer to an in-video question of video $v$, we assume the eLearning platform will record the student's answer, the points rewarded $o_{uv}$, and the maximum possible points of this question $o_{v}^{max}$. We will consider student performance on their first attempt at the quiz, i.e., whether they are Correct on First Attempt (CFA) or not (non-CFA), as a measurement of their knowledge acquisition from the video~\cite{Brinton2016MiningMC}. Therefore, we formulate the CFA score for student $u$ on lecture video $v$ as $\mathsf{CFA}_{uv} = 1$ when $o_{uv} = o_{v}^{max} $; otherwise, $\mathsf{CFA}_{uv} = 0$. 

We consider time-varying encodings and static encodings of clickstream logs for different purposes.
The \emph{time-varying encoding} $F_{uv} = \{E_{uv}(1), ..., E_{uv}(i),..., E_{uv}(L_f)\}$ is the sequence containing the entire event vector $E_{uv}(i)$ prior to the first time a user $u$ answered the question for video $v$, where $L_f \leq L_{uv}$ represents the number of clicks immediately before the student submits the answer.
The time-varying encoding will be used in Section~\ref{sec:cfa} for modeling student video-watching behavior in the CFA prediction algorithm.
On the other hand, the \emph{static encoding} $S_{uv}$ serves as a time-independent representation that summarizes the clickstreams after student $u$ finishing watching video $v$. We count two types of information for $S_{uv}$: (i) the number of total clicks $L_{uv}$ and (ii) the number of each event type the student has made by the end of the video.
The static encodings will be used in Section~\ref{sec:optim} for clustering students into groups.

\subsection{Self-Supervised Clickstream Pre-training}
\label{sec:pre}
We first aim to develop a generative model of how users make clicks on videos.
Doing so can provide a denoised representation of the student video-watching process for CFA prediction.
A key challenge we face is that we do not have a large amount of historical clickstreams data that can help us model the sequential information in our setting compared with other learning applications, e.g., Natural Language Processing (NLP). 
To overcome this, we propose a method for learning students' sequential clicking behavior from our extracted dataset based on a surrogate objective.
Specifically, we design a pre-training method for our model based on the concept of Continuous Bag Of Words (CBOW)~\cite{Mikolov2013EfficientEO}, which has achieved notable success in NLP community. 
In our setting, the distributed representations of surrounding clicks are combined to predict the current click. To the best of our knowledge, no other clickstream research has leveraged pre-training techniques to provide prediction models better data understanding. 

\begin{figure}[t]
\centering
\includegraphics[width=1\linewidth]{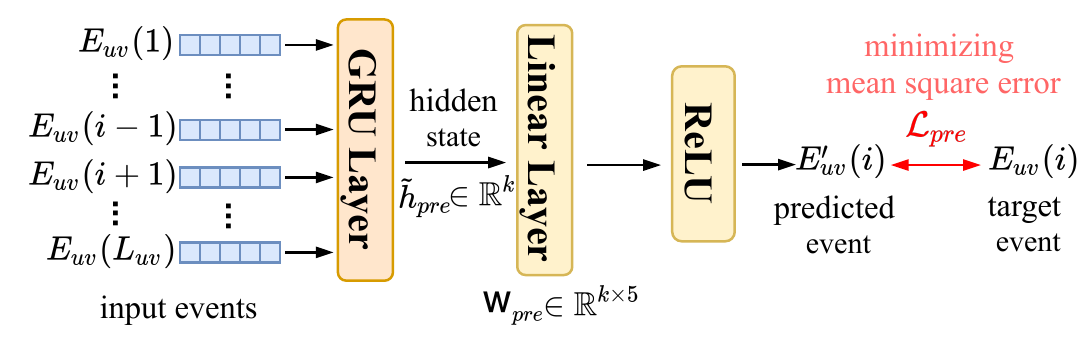}
\caption{The self-supervised clickstream pre-training architecture. The model learns to predict the $i$-th event $E_{uv}(i)$ made by user $u$ on video $v$, from surrounding clicks. The pre-training objective minimizes the mean square error $\mathcal{L}_{pre}$ between the predicted event and ground-truth event.} 
\label{fig:pretrain}
\vspace{-0.1in}
\end{figure}

Formally, for each student-video pair, we extract each click $E_{uv}(i) \in E_{uv}$ as a target and train the model by taking the rest of the clicks $E_{uv} \setminus E_{uv}(i)$ as input. 
We leverage Gated Recurrent Units (GRU) as our learning model, which we will formalize in detail in Section~\ref{sec:cfa}.
Therefore, the size of this pre-training dataset is $\sum_{u,v} L_{uv}$ which is larger than the summation of individual user-video pairs. 
The resulting pre-training architecture is shown in Fig.~\ref{fig:pretrain}.
The GRU outputs the final hidden state $\widetilde{h}_{pre}$ by taking all event vectors except for $E_{uv}(i)$ as input.
The linear layer then output the predicted event $E'_{uv}(i) \in \mathbb{R}^{5}$, which is of the same size as $E_{uv}(i)$, and is subsequently passed through a rectified unit (ReLU):
\begin{equation} 
    E'_{uv}(i) = \textup{ReLU}( \textbf{W}_{pre}^T \cdot \widetilde{h}_{pre} + \textbf{b}_{pre} ),\\
\end{equation}
where $\textbf{W}_{pre} \in \mathbb{R}^{k \times 5}$ is the weight matrix for $\widetilde{h}_{pre}$, $k$ is hidden dimension, and $\textbf{b}_{pre} \in \mathbb{R}^{5}$ is the bias vector.
For training, we minimize the mean square error $\mathcal{L}_{pre}$ between the predicted event $E'_{uv}(i) \in \mathbb{R}^{5}$ and the target event $E_{uv}(i)$:
\begin{equation} 
    \mathcal{L}_{pre} = \sum_{(u,v) \in \mathcal{D}} \frac{1}{L_{uv}} \sum_{i=1}^{L_{uv}}(E_{uv}(i) - E'_{uv}(i))^2.\\
\end{equation}

\subsection{CFA Prediction Model}
\label{sec:cfa}
To model the impact of students' time-dependent clicking behavior on their in-video knowledge acquisition, we use the time-varying encoding vector $F_{uv}$ as the input, and the score $\mathsf{CFA}_{uv}$ as the prediction target.
We leverage GRU, known for its ability to capture dependencies over long time periods from time-series data~\cite{Chung2014EmpiricalEO}, as our CFA prediction model. The objective of the model is to generate a learned representation from raw data for each time-varying feature.
The hidden state of the model is formulated as:
\begin{equation} 
    h_i = \textup{GRU}(E_{uv}(i), h_{(i - 1)}).
\end{equation}
The hyperparameters of GRU network are initialized through the pre-training process in Section~\ref{sec:pre}.
We will see in Section~\ref{sec:experiment} how leveraging the pre-training results leads to significant improvement in CFA prediction performance.

The final state $h_{L_{f}}$ at the last time index $L_{f}$ is taken as GRU's output and serves as the click representation for user $u$ watching video $v$.
A linear layer then transforms the final hidden state $h_{L_{f}}$ into predicted CFA score $C'_{uv} \in \mathbb{R}^{2}$:
\begin{equation} 
    C'_{uv} = \textup{softmax}(\textbf{W}_{C}^T \cdot h_{L_{f}} + \textbf{b}_{C}),\\
\end{equation}
where $\textbf{W}_{C} \in \mathbb{R}^{k \times 2}$ is the weight matrix for $h_{L_{f}}$, $k$ is hidden dimension, and $\textbf{b}_{C} \in \mathbb{R}^{2}$ is the bias vector.
Note that the linear layer is distinct from the pre-training linear layer in Section~\ref{sec:pre};
compared with the main GRU layer that has pre-trained weights for its initialization, this linear layer is trained from scratch.
The model can then be optimized by minimizing the binary cross entropy (BCE) loss $\mathcal{L}_{CFA}$:
\begin{equation} 
    \mathcal{L}_{CFA} = -\sum_{(u,v) \in \mathcal{D}} C_{uv}^T  \log(C_{uv}') + (\textbf{1} - C_{uv})^T  \log( \textbf{1} - C_{uv}'),\\
\label{eq:BCEloss}
\end{equation}
where \textbf{1} is an all-one vector and $C_{uv} \in \{0,1\}^2$ is the one-hot encoding vector of $\mathsf{CFA}_{uv}$ (i.e., if $\mathsf{CFA}_{uv} = 1$ then $C_{uv} = (1,0)^T$, else $C_{uv} = (0,1)^T$).

\subsection{Clustering-Guided Meta-Learning}
\label{sec:optim}



We develop a clustering methodology based on the static encoding vector $S_{uv}$ to extract similar clicking behavior between students and separate students into groups. 
Based on these clusters, we introduce a novel meta-learning procedure that guides our model training process to reflect students similarities for CFA prediction optimization.

Consider our CFA prediction problem with training dataset $\textbf{D}_{train} = \{\textbf{x}_n^{train}, \textbf{y}_n^{train}\}_{n=1}^{N}$, where $\textbf{x}_n^{train}$ denotes the $n$-th sample (i.e., $E_{uv}$ for a student-video pair) and $\textbf{y}_n^{train} \in \{0,1\}^c$ is the class label vector with $c$ classes ($c = 2$ in our case).
We aim to obtain the optimal CFA prediction model parameter vector $\textbf{w}^{*}$ by minimizing the BCE loss from Equation (\ref{eq:BCEloss}) on the training set as:
\begin{equation} 
\begin{split}
    &\textbf{w}^{*} = \operatorname*{argmin}_\textbf{w} \mathcal{L}_{CFA} \\
    &= \operatorname*{argmin}_\textbf{w} \frac{1}{N} \sum_{n=1}^{N} \ell_{CFA}(\textbf{y}_n^{train}, f(\textbf{x}_n^{train}, \textbf{w})),\\
\end{split}
\end{equation}
where $\ell_{CFA}$ is the BCE loss for a particular sample, and $f$ is the CFA predictor.
We denote $\ell_{CFA}(\textbf{y}_n^{train}, f(\textbf{x}_n^{train}, \textbf{w}))$ as $\mathcal{L}_{n}^{train}(\textbf{w})$ for notational convenience.
For meta-learning, we introduce a weighting function $\mathcal{W}(\mathcal{L}_{n}^{train}(\textbf{w}); \Theta)$ on the $n$-th sample loss in order to enhance the training process, where $\Theta$ denotes the parameters of weighting network $\mathcal{W}(\cdot)$.
With sample re-weighting based on $\mathcal{W}$, the optimal parameter $\textbf{w}^{*}$ can be expressed as:
\begin{equation} 
\begin{split}
    &\textbf{w}^{*}(\Theta) \triangleq \operatorname*{argmin}_\textbf{w} \frac{1}{N} \sum_{n=1}^{N} \mathcal{W}(\mathcal{L}_{n}^{train}(\textbf{w}); \Theta) \mathcal{L}_{n}^{train}(\textbf{w}).\\
\end{split}
\label{eq:classifier}
\end{equation}

Traditional meta-learning research employs a clean meta(or validation)-dataset $\textbf{D}_{meta} = \{\textbf{x}_m^{meta}, \textbf{y}_m^{meta}\}_{m=1}^{M} (M \ll N)$ to solve the label-imbalance issue. In our setting, we aim to guide our classification model to absorb and generalize information from clusters of student clicking behaviors. Thus, we employ the static encoding $S_{uv}$ as a clustering criteria and apply $k$-means~\cite{Hartigan1979AKC} on the meta-dataset $\textbf{D}_{meta}$ to generate a clustering-guided meta-dataset:
\begin{equation} 
\begin{split}
    &\mathbf{D}_{meta} = \{\textbf{D}_{meta}^{1},...,\textbf{D}_{meta}^{N_c}\} = \texttt{k-means} (\textbf{D}_{meta} | S_{uv}),\\
\end{split}
\label{eq:kmean}
\end{equation}
where $\textbf{D}_{meta}^{p} = \{\textbf{x}_m^{meta}, \textbf{y}_m^{meta}\}_{m=1}^{K_p}$ denotes the sample set of the $p$-th cluster and $K_p$ represents the number of samples in the $p$-th cluster.
The number of clusters $N_c$ is decided by the highest silhouette score on the clustered datapoints $\textbf{D}_{meta}$.
We order $\textbf{D}_{meta}^{p}$ by the entropy of labels (from lowest to highest) in each set of datapoints.

The question now is how to define the weighting function $\mathcal{W}$. Intuitively, the key idea is to map from training loss to sample weights.
This process will iterate between (i) updating weights based on the current training loss values and (ii) minimizing the weighted training loss for classifier updating. 
Traditional research in meta-learning has employed a fixed weighting function; we instead base our procedure on MW-net~\cite{Shu2019MetaWeightNetLA} which introduces a data-driven, dynamic weighting function that automatically learns the hyperparameter set $\Theta$.
In doing so, instead of iterating across the entire meta-dataset $\textbf{D}_{meta}$ for optimizing $T$ epochs, we use each cluster $\textbf{D}_{meta}^p$ for $T / N_c$ epochs. 
That is, we divide the training process by the number of clusters and use meta-learning on each student behavioral cluster for tuning based on information in each group. 
Formally, the hyperparameter set $\Theta$ of the weighting function $\mathcal{W}(\mathcal{L}_{n}^{train}(\textbf{w}); \Theta)$ can be optimized by:
\begin{equation} 
    \Theta^{*} = \operatorname*{argmin}_\textbf{$\Theta$} \frac{1}{M} \sum_{p=1}^{N_c} \sum_{m = 1}^{K_p} \mathcal{L}_{m}^{meta, p}(\textbf{w}^{*}(\Theta)),
\label{eq:metanet}
\end{equation}
where
$$\mathcal{L}_{p}^{meta, p}(\textbf{w}^{*}(\Theta)) = \ell_{meta}(\textbf{y}_m^{meta}, f(\textbf{x}_m^{meta}, \textbf{w}))$$
for $m = 1,...,K_p$ represents the loss between sample and label for $p$-th cluster, and $\ell_{meta}(\cdot)$ is the loss function of the weighting network. 

The nested structure in~\eqref{eq:metanet} specifies that the optimal $\Theta^{*}$ is based on the loss of each cluster $\mathcal{L}^{meta, p}$ at the optimal CFA prediction model $\textbf{w}^{*}$, which in turn is dependent on $\Theta$ and $\mathcal{L}^{train}$.
The gradient descent learning process of the classifier network~\eqref{eq:classifier} at time step $t$ is formulated as:
\begin{equation} 
\begin{split}
    &\widehat{\textbf{w}}^{t}(\Theta) = \textbf{w}^t - \alpha \frac{1}{|\mathcal{B}_t|} \sum_{n \in \mathcal{B}_t} \mathcal{W}(\mathcal{L}_{n}^{train}(\textbf{w}^t); \Theta) \nabla_{\textbf{w}} \mathcal{L}_{n}^{train}(\textbf{w}^t)) \bigg\rvert_{\textbf{w}^t},
\end{split}
\label{eq:manner}
\end{equation}
where $\mathcal{B}_t$ is the mini-batch for the training set at time $t$ and $\alpha$ is the step size of the SGD optimizer.
The gradient step for $\Theta_{p}^t$ toward $\Theta^{*}$ based on the loss of~\eqref{eq:metanet} on the $p$-th clustering-guided meta-dataset can be calculated as:
\begin{equation} 
\begin{split}
    &\Theta_{p}^{t+1} = \Theta_{p}^{t} - \beta \frac{1}{|\mathcal{B}^p_t|} \sum_{m \in \mathcal{B}^p_t} \nabla_{\Theta} \mathcal{L}_{m}^{meta, p}(\widehat{\textbf{w}}^{t}(\Theta))\bigg\rvert_{\Theta_{p}^t},
\end{split}
\label{eq:updatetheta}
\end{equation}
where $|\mathcal{B}^p_t|$ is the mini-batch for meta-cluster $p$ at time $t$ and $\beta$ is the step size.
By taking a gradient descent step towards the optimal $\textbf{w}^{*}$ based on the updated $\Theta_{p}^{t+1}$, the updated parameter vector of the CFA classifier can be written as:
\begin{equation} 
\begin{split}
    &\textbf{w}^{t+1} =\\ &\textbf{w}^t - \alpha \frac{1}{|\mathcal{B}_t|} \sum_{n \in \mathcal{B}_t} \mathcal{W}(\mathcal{L}_{n}^{train}(\textbf{w}^t); \Theta_{p}^{t+1}) \nabla_{\textbf{w}} \mathcal{L}_{n}^{train}(\textbf{w}^t)) \bigg\rvert_{\textbf{w}^t}.
\end{split}
\label{eq:updateW}
\end{equation}
For the weighting network $\mathcal{W}$, we select a Multi-Layer Perceptron (MLP) network with a single hidden layer containing 100 hidden nodes.
Given the loss value $\mathcal{L}_{i}^{train}(\textbf{w})$ as input, the weighting function outputs a weighting value for modifying the training loss.
The overall meta-learning optimization procedure we have developed here is summarized in Algorithm~\ref{tb:optimization}.

\begin{algorithm}[t]
\caption{Clustering guided meta-learning-based optimization for CFA prediction}
\begin{algorithmic}[1]
\State \textbf{Input}: Training dataset $\textbf{D}_{train}$ and meta-dataset $\textbf{D}_{meta}$
\State Clustered-meta-dataset $\textbf{D}_{meta}^{p} = \texttt{k-means}(\textbf{D}_{meta}| S_{uv})$
\State Initializing CFA prediction model parameter $\textbf{w}^0$ and weighting model parameter $\Theta^0$
\For{ cluster $ p \gets 1$ to $N_c$}
\State Formulate the CFA model parameter $\widehat{\textbf{w}}^{t}(\Theta)$ by~\eqref{eq:manner}
\For{ epoch $\gets 1$ to $T / N_c$}
\State $\nabla_{\Theta} \gets$ \textbf{BackProp}($\mathcal{L}^{meta, p}$, $\Theta_p^t$)
\State $\Theta_p^{t+1} \gets$ \textbf{OptimizerStep}($\Theta_p^{t}$, $\nabla_{\Theta}$) (Eq.~\ref{eq:updatetheta})
\State $\nabla_\textbf{w} \gets$ \textbf{BackProp}($\mathcal{L}^{train}$, $\textbf{w}^t$)
\State $\textbf{w}^{t+1}$ $\gets$ \textbf{OptimizerStep}($\textbf{w}^{t}$, $\nabla_\textbf{w}$) (Eq.~\ref{eq:updateW})
\EndFor
\EndFor
\end{algorithmic}
\label{tb:optimization}
\end{algorithm}

\section{Experimental Evaluation and Analytics}
\label{sec:experiment}
We now conduct experiments to evaluate our CFA prediction methodology from Section~\ref{sec:method}. After describing our datasets (Section~\ref{sec:dataset}), baseline predictors (Section~\ref{sec:baseline}), and experimental setup (Section~\ref{sec:setup}), we will present our prediction results (Section~\ref{sec:qualityanalysis}) and discuss learning and content analytics that arise from our model (Section~\ref{ssec:analytics}). Finally, we will conduct ablation studies to assess the importance of different components of our methodology (Section~\ref{sec:ablation}).

\subsection{Datasets}
\label{sec:dataset}

\begin{table*}
\begin{center}
\scalebox{1.05}{
\begin{tabular}{lccccccc}
\hline
   dataset  & lecture videos  & video length (sec) & quizzes & users & clickstream events & user-video pairs & avg. CFA score \\
 \hline\hline
edX-Purdue & 2,663 & 385 & 649 & 15,133 & 1,001,727 & 150,378 & 0.64 \\

Coursera-FMB & 92 & 1,015 & 92 & 3,770 & 832,342 & 38,695 & 0.65   \\
Coursera-NI & 115  & 327  & 69 & 2,680 & 753,833 & 44,537 & 0.74  \\
\hline
\end{tabular}}
\end{center}
\caption{Basic information on our edX and Coursera eLearning datasets used in our evaluation.}
\label{tb:dataset}
\vspace{-0.15in}
\end{table*}
We leverage online learners' clickstreams from two video-based eLearning platforms, edX and Coursera. Both platforms log users' interaction with the video player based on the format specified in Section~\ref{sec:extr}. 
\begin{itemize}
    \item From the edX platform, we consider 51 graduate-level online certificate courses from Purdue University. We denote the collection of these courses as ``edX-Purdue''\footnote{edX-Purdue: www.edx.org/masters/online-masters-in-electrical-and-computer-engineeringmasters-lead-capture-form}. Each of these courses covers topics related to electrical engineering and computer science, and was collected since 2019.
    \item For the Coursera platform, the datasets correspond to two offerings of networking-related massive open online courses (MOOCs) taught by the authors: Networks: Friends, Money, and Bytes\footnote{FMB: www.coursera.org/course/friendsmoneybytes} (``Coursera-FMB'') and Networks Illustrated: Principles Without Calculus\footnote{NI: www.coursera.org/course/ni} (``Coursera-NI'').
\end{itemize}

\begin{table*}
\begin{center}
\scalebox{1.05}{
\begin{tabular}{ccccccc}

\hline
  dataset & \multicolumn{2}{c|}{edX-Purdue} & \multicolumn{2}{c|}{Coursera-FMB} & \multicolumn{2}{c}{Coursera-NI}\\
\hline\hline
 & ACC & F1 & ACC & F1 & ACC & F1 \\
\hline
3-gram & .6723 $\pm$ .0214 & .6103 $\pm$ .0223 & .6598 $\pm$ .0137 & .5764 $\pm$ .0268 & .6629 $\pm$ .0203  & .5837 $\pm$ .0107\\

4-gram & .6749 $\pm$ .0229 & .6092 $\pm$ .0397 & .6617 $\pm$ .0203 & .5920 $\pm$ .0202 & .6693 $\pm$ .0231 & .5866 $\pm$ .0431\\

latent-var & .6849 $\pm$ .0132 & .6232 $\pm$ .0282  & .6783 $\pm$ .0092 & .6018 $\pm$ .0137 & .6820 $\pm$ .0089 & .6018 $\pm$ .0161 \\

CNN & .6731 $\pm$.0082 & .6172 $\pm$ .0379 & .6639 $\pm$ .0132 & .5932 $\pm$ .0261 & .6791 $\pm$ .0106 & .5973 $\pm$ .0209\\

GRU & .6861 $\pm$ .0121 & .6115 $\pm$ .0220 & .6739 $\pm$ .0104 & .6022 $\pm$ .0379 & .6825 $\pm$ .0098 & .6017 $\pm$ .0185\\
pre-GRU & .7026 $\pm$ .0117 & .6329 $\pm$ .0353 & .6813 $\pm$ .0129 & .6178 $\pm$ .0206 & .6931 $\pm$ .0108 & .6095 $\pm$ .0195 \\

GRU-meta(C1) & .7186 $\pm$ .0138 & .6382 $\pm$ .0311 & .6979 $\pm$ .0113 & .6293 $\pm$ .0367 & .7058 $\pm$ .0120 & .6197 $\pm$ .0233\\

GRU-meta(C2) & .7209 $\pm$ .0127 & .6401 $\pm$ .0124 & .7037 $\pm$ .0121 & .6261 $\pm$ .0204 & .7002 $\pm$ .0114 & .6132 $\pm$ .0179\\

pre-GRU-meta(C1) & .7281 $\pm$ .0114 & .6503 $\pm$ .0109 & .7029 $\pm$ .0101 & .6237 $\pm$ .0133 & .7136 $\pm$ .0122 & .6209 $\pm$ .0106\\

pre-GRU-meta(C2) & \textbf{.7298 $\pm$.0125} & \textbf{.6567 $\pm$ .0173} & \textbf{.7142 $\pm$.0110} & \textbf{.6308 $\pm$ .0198} & \textbf{.7139 $\pm$ .0129} & \textbf{.6216 $\pm$ .0181}\\
\hline

\end{tabular}}
\end{center}
\caption{Prediction results obtained for each baseline and our methododlogy in terms of classification accuracy (ACC) and F1 score. 3/4-gram and latent-var denote the $n$-gram model and behavior-based latent variable model, respectively. Overall, we see that our methodology significantly outperforms the baselines for each metric on each dataset, and that the pre-training and meta-learning components both improve performance.}
\label{tb:performance}
\vspace{-0.25in}
\end{table*}

Basic statistics of our three datasets are given in Table~\ref{tb:dataset}.
For each dataset, lecture students are paired with their in-video quiz submissions, resulting in 150,378, 38,695, and 44,537 user-video-pairs for edX-Purdue, Coursera-FMB, and Coursera-NI, respectively.
Comparing each dataset, edX-Purdue and Coursera-NI have more and shorter-length (327 and 385 seconds on average) lecture videos, while Coursera-FMB has fewer and longer-length lecture videos (1,015 seconds).
The fraction of lecture videos that have quizzes is higher for Coursera-FMB (1.0) and Coursera-NI (0.6) is higher than edX-Purdue (0.2). Still, the 51 certificate courses of the edX-Purdue dataset provide sufficient user-video pairs for analysis.

\subsection{Baselines}
\label{sec:baseline}
We consider two existing behavior-based CFA prediction methods as baselines, as well as several variations/configurations of our proposed methodology:

\subsubsection{Behavior-based latent variable model}
The method in~\cite{Lan2017BehaviorBasedLV} introduces nine hand-crafted behavioral features to summarize a learner's consumption of a lecture video, including the fraction of the video the learner completed, the number of play and pause events, and the standard deviation of playback rate.
The coupling of a learning model -- which include a latent variable definition of engagement based on these behaviors -- with a response model is used to predict the CFA score.
We denote this baseline as ``latent-var''. 

\subsubsection{$n$-gram model}
\cite{Jeon2020DropoutPO_1} proposed a binary classification model that encodes clicking events as $n$-gram vectors of event types (i.e., the $e_i$ attribute in Section~\ref{sec:extr}-1) and uses them as input to GRUs. 
The authors found that $n$-gram event vectors are better representations of a student's behavior than single events in the sense of improving prediction quality.
They find $n=4$ to have the best performance, In our experiment, we use two variants, $n=3$ and $n=4$, and denote these baselines as ``3-gram'' and ``4-gram'', respectively.

\subsubsection{Variants of our method}
While our approach in Section~\ref{sec:method} is based on RNNs, other model structures can also be employed.
For example, instead of using GRUs, we will also experiment with using a one-dimensional convolutional neural network (CNN)~\cite{Hu2014ConvolutionalNN} to model students' video-watching behavior. 
This comparison enables us to validate the effectiveness of GRUs in terms of representing sequences of clickstream events compared to other neural network structures. 
For this, we implement a one-dimensional CNN with ReLU activation function for the architecture in Section~\ref{sec:cfa}.

\subsubsection{Configurations of our method}
To evaluate the quality of the proposed clickstream pre-training (Section~\ref{sec:pre}) and clustering-based meta-learning (Section~\ref{sec:optim}) methods , we also compare several different configurations of our methodology in our experiments.
We include ``pre" in the algorithm title to denote pre-training is enabled, and ``meta" to denote meta-learning is enabled.
Recall that the static encoding $S_{uv}$ we defined for meta-learning includes two components: (1) the total number of clicks and (2) the vector containing the number of each event type. We add ``(C1)'' and ``(C2)'' to the meta-learning algorithm title.

\begin{figure*}[t]
\centering
\includegraphics[width=1.0\linewidth]{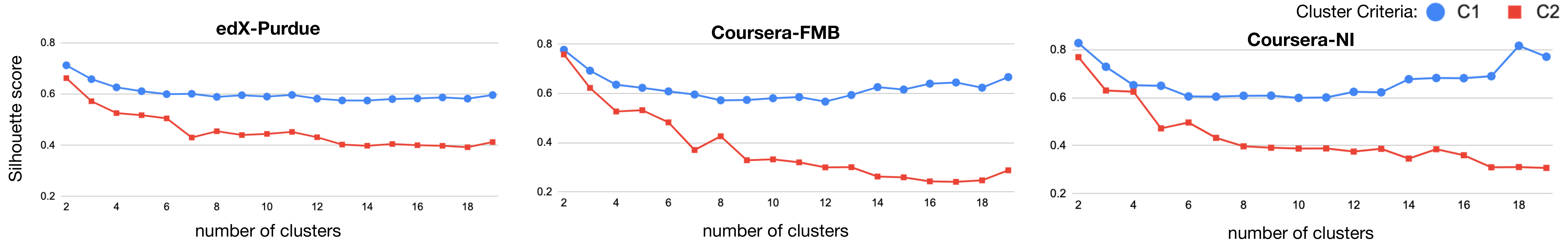}
\caption{Plots of the silhouette score for the meta-dataset when $k$-means separates it using 2 to 19 clusters on the three datasets.} 
\label{fig:clusterscore}
\vspace{-0.1in}
\end{figure*}

\subsection{Experimental Setup and Evaluation Metrics}
\label{sec:setup}
We randomly partition student-video pairs into 5 equally-sized data folds and perform 5-fold cross-validation on each dataset.
Both the training (e.g., clickstream pertaining, CFA model, and clustering-guided meta-learning) and the testing processes are performed individually on each dataset.
The latent embedding dimension in our model is set to $k = 128$.
During training, the batch size is set to $32$, the learning rate to $0.001$, and the maximum training epochs to $100$.
These settings apply to all methods; we found the performance of these methods is stable across a wide range of values for these parameters. 
To evaluate the performance of each method, we apply two commonly used classification metrics: prediction accuracy (ACC) and F1 score.
ACC measures the average accuracy of CFA prediction on the test set, i.e., the number of correctly classified data instances divided by the total number of data instances.
The F1 score is the harmonic mean of precision and recall.

\begin{figure*}[t]
\centering
\includegraphics[width=\linewidth]{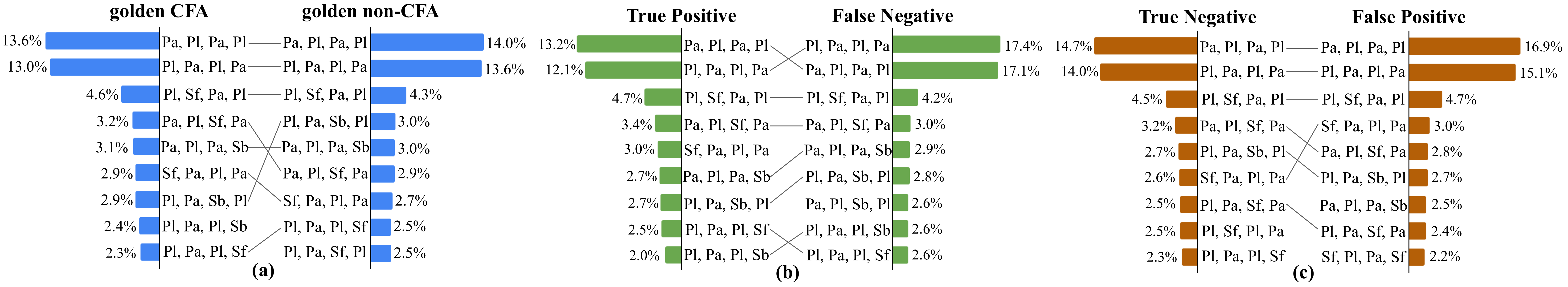}
\caption{Illustration of behavioral analytics that can be derived from our model. (a) Plot of the 4-gram sequential event distribution for CFA student-video pairs and non-CFA student-video pairs across three test sets. By separating the test sets based on the correctness of pre-GRU-meta(C2)'s prediction, we also show the sequential event distribution of the samples of (b) true positive and false negative as well as (c) true negative and false positive. The abbreviations Pl, Pa, Sb, Sf, and Sp represent the event type of play, pause, skip back, skip forward, and speed change, respectively.}
\label{fig:freq}
\vspace{-0.2in}
\end{figure*}

\subsection{Overall Predictive Quality}
\label{sec:qualityanalysis}
The overall performance obtained by our methodology and the baselines is summarized in Table.~\ref{tb:performance}.
In the top five rows of the table, we first compare three types of input features, without pre-training or meta-learning: (1) the combination of sequential event types used for $n$-gram models, (2) nine engagement-related factors used for latent-var, and (3) the raw time-varying feature $F_{uv}$ used in our setting (note that we include two basic models, CNN and GRU, in this comparison).
These five models perform similarly in terms of predictive accuracy, but the higher variance of $n$-gram models shows that the raw features are more sensitive to hyperparameter values than the hand-crafted features used in latent-var and the raw inputs used in CNN and GRU.
The model structure of latent-var based on carefully crafted behavioral features results in slightly higher accuracy than CNN and GRU; however, the proposed method takes in less data (five raw inputs at each time step) than latent-var (nine features at each time step), without needing any feature engineering.
Taking the same input ($F_{uv}$), the accuracy of GRU is higher than that of CNN since the sequential nature of clickstream data is more suitable for recurrent neural networks (GRU), which is effective at capturing data dynamics.

For meta-learning, we perform $k$-means algorithm to cluster $\textbf{D}_{meta}$ into $\textbf{D}_{meta}^{p}$, and we first analyze the number of clusters $N_c$ separately for each meta-dataset.
Fig.~\ref{fig:clusterscore} shows the silhouette score for different numbers of clusters and clustering criteria on our three datasets.
For each meta-dataset, we identify that the best number of clusters is 2.
With this setting, in our methodology, each cluster is used half of the time during the optimization process. During the CFA model's training process, the cluster with lower entropy is used in the first $\lceil T/2 \rceil$ epochs while the cluster with higher entropy is used in the rest of the epochs.

Based on this, we compare the performance of all baselines and our full framework―that is, pre-GRU-meta(C1) and pre-GRU-meta(C2).
By tuning the training loss of the CFA model based on the weighting function, the model's optimization direction is guided by each meta-data cluster and reflects the data distribution within each group. 
Table~\ref{tb:performance} shows the predictive accuracy of all methods; we see that our proposed method consistently achieves higher accuracy than all baseline models on both metrics.
Moreover, these results show that bringing the similarity within student behavioral clusters can indeed benefit the predictive accuracy of the trained model.
Our model achieves its biggest quality improvement when clustering based on C2, with an average gain of $5.06\%$ in ACC over the $4$-gram model and $3.76\%$ over the latent-var model. 

\begin{figure*}[t]
\centering
\includegraphics[width=\linewidth]{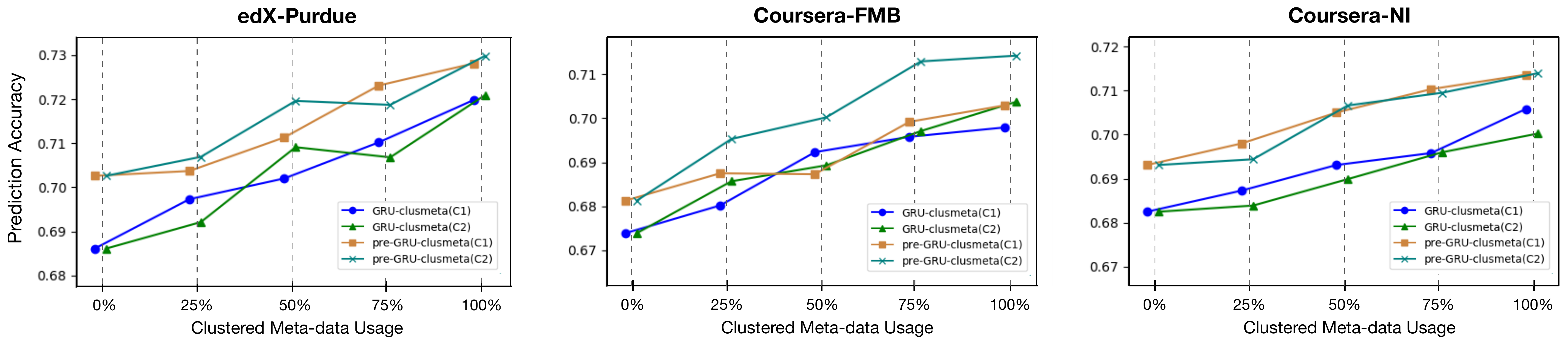}
\caption{Average prediction accuracy obtained with different fractions of meta-dataset usage on the three datasets. The $x$-axis represents the proportion of meta-dataset usage, where $0\%$ means that the model training does not incorporate meta-learning and $100\%$ means the meta-dataset is fully used to optimize the training process. The increasing trend of the prediction accuracy in each case shows the efficacy of proposed meta-learning-based training procedure.} 
\label{fig:ablation}
\vspace{-0.18in}
\end{figure*}

\subsection{Model Analytics}
\label{ssec:analytics}
We also explore potential learning and content analytics that can be derived from our model.
In Fig.~\ref{fig:freq}, we qualitatively analyze the most frequently occurring $4$-gram event patterns based on our best model (pre-GRU-meta(C2)).
We consider the differences in distributions between the true CFA and non-CFA sequences in Figure~\ref{fig:freq}(a), and between true positive (TP), false negative (FN), true negative (TN), and false positive (FP) in Figure~\ref{fig:freq}(b) and (c).
The order of the top three sets of $4$-gram patterns are similar for each of the sample distributions, but they are also the dominant click patterns in the test sets.
We find that our model has a tendency to predict more accurately on sequences that exhibit more varied patterns: the FN and FP samples have an average of $33\%$ $4$-grams that belong to the top two most frequent sequences, while this percentage for TN and TP is $27\%$.
It would be difficult for an instructor to manually evaluate students' performance based on continuous play and pause events in isolation since the actual status of the student is unknown when they make these repeated clicks (e.g., the students might either be focusing on the video content or being inattentive and impatient).
Thus, providing the predictions (indicators of knowledge acquisition) together with the observed sequences (indicators of learning styles) can be useful for instructors.

We also find that sequences containing skip forward (Sf) are highly predictive (e.g., Sf appears in $9$ of the top sequences in the TP and TNs).
For example, the model performs better on sequences containing “Sf, Pa, Pl, Pa” and “Pl, Pa, Pl, Sf”.
In more detail, there are five typical sequences with Sf in the TN set, which shows that our model correctly predicts the performance of non-CFA students when they skip forward.
These results show that our model's behavior incorporates components of human judgment, specifically in the case that students who frequently skip certain parts of video content often perform poorly on the in-video quizzes.


\subsection{Ablation Studies}
\label{sec:ablation}
We conduct ablation studies to verify the importance of three components in our methodology:

\subsubsection{Clickstream pre-training}
To analyze the performance gain as a result of our self-supervised clickstream pre-training method, we compare GRU and pre-GRU (i.e., without meta-learning) as well as GRU-meta and pre-GRU-meta (i.e., with meta-learning). 
The pre-trained models allow the CFA training process to start with better-initialized event embeddings that capture the nature of clickstream sequences identified through this pre-training. Therefore, fine-tuning on the CFA scores leads to consistently better performance on all the datasets.
In Table~\ref{tb:performance}, the models with clickstream pre-training outperform models without pre-training by about $1.5\%$ in ACC.

\subsubsection{Clustering-guided meta-learning}
To evaluate the effectiveness of clustering-guided meta-learning, we compare GRU against GRU-meta (i.e., without pre-training) and also compare pre-GRU against pre-GRU-meta (i.e., with pre-training).
In Table~\ref{tb:performance}, we find that models with clustering-guided meta-learning applied during trained outperform models without this training method in ACC by about $2.8\%$ on average.
These results show that optimizing the prediction model based on meta-learning to absorb clustering information from small samples can improve the performance.
In general, the performance of the models guided by C2 are slightly higher than that by C1, showing that using more specific criteria for separating students into groups (namely, specific event types) can result in higher performance.
Note, however, that C2 consistently outperforms C1 only under the model with per-training enabled, further emphasizing the benefit provided by our holistic methodology.

\subsubsection{Meta-data usage}
To further analyze the performance of clustering-guided meta-learning, we construct ablation studies by modifying the meta-dataset usage.
At each time, we randomly select a quarter of the meta-data, leave them out, and then perform $k$-means clustering and meta-learning optimization based on the rest of the data. 
We conduct this experiment for all the models that use clustering-guided meta-learning during training.
Fig.~\ref{fig:ablation} shows the average prediction accuracies of different models' structures under different proportions of meta-data usage.
We plot the performance of the models from without using meta-learning entirely ($0\%$) to using the full meta-dataset ($100\%$) during the optimization process.
In each case, we observe a growing trend for prediction accuracy when a larger proportion of the meta-dataset is used.
These results show that the CFA model performs better when more information from meta-data is used for tuning the weighting function and the loss of the CFA model.

\section{Conclusion and Future Work}
We developed a novel methodology for predicting student performance on in-video quizzes from their associated video-watching behavior. Rather than relying on hand-crafted feature engineering, we model student video-watching behavior through deep learning operating on raw event data. To initialize our prediction model, we developed a self-supervised clickstream pre-training method that functions as a generative model for student clicks. Moreover, we developed a novel clustering guided meta-learning-based training procedure that optimizes the prediction model based on inferred similarities within student behavioral clusters. Our subsequent evaluation on three real-world eLearning datasets confirmed that our methodology achieves superior quality in predicting students' performance compared to baseline prediction algorithms.
Moreover, we showed that our model's interpretation of clicks tends to match that of human judgment, and that the sequence-CFA relationships learned by our methodology provide useful analytics on student behavior.

The eLearning datasets employed in this work cannot be made publicly available due to data privacy concerns. Thus, an immediate step we are pursuing is identifying a public dataset which can be used by the community for reproducibility of our results. Additionally, our focus here has been on modeling the in-video clickstream data. eLearning platforms also contain an abundance of data sources from other learning modalities, e.g., question text and forum discussions. This data can provide an additional layer to understanding the relationships between student behavior and performance in eLearning courses. Therefore, one direction of future work includes integrating these other sources into a more holistic prediction model.

While we presented our methodology in this paper based on the educational setting, future work can investigate how it can be generalized to other online video-watching settings. On video sharing/recommendation platforms like YouTube, for example, it may be possible to adapt this methodology to predicting video preferences based on a viewer's click behavior, with meta learning guided by clusters of how users consume content. 

\section*{Acknowledgements}
Y.\ Chu, L.\ Cruz, K.\ Douglas, and C.\ Brinton were supported in part by the Charles Koch Foundation.
A.\ S.\ Lan was supported in part by the National Science Foundation under grant IIS-1917713. 


\bibliographystyle{IEEEtran} 
\bibliography{IEEEbib}

\vspace{12pt}

\end{document}